\documentclass[a4paper]{article}

\usepackage{INTERSPEECH2022}
\usepackage{tipa}
\usepackage{url}

\usepackage{breakurl}
\usepackage[breaklinks,hidelinks]{hyperref}

\title{Modeling speech recognition and synthesis simultaneously: Encoding and decoding lexical and sublexical semantic information into speech with no direct access to speech data}
\name{Ga\v{s}per Begu\v{s}$^1$, Alan Zhou$^1$}
\address{
  $^1$University of California, Berkeley
  }
\email{begus@berkeley.edu, azhou314@berkeley.edu}

\begin{document}

\maketitle
\begin{abstract}
Human speakers encode information into raw speech which is then decoded by the listeners. This complex relationship between encoding (production) and decoding (perception) is often modeled separately. Here, we test how encoding and decoding of lexical semantic information can emerge automatically from raw speech in unsupervised generative deep convolutional networks that combine the production and perception principles of speech. We introduce, to our knowledge, the most challenging objective in unsupervised lexical learning: a network that must learn unique representations for lexical items with no direct access to training data. We train several models (ciwGAN and fiwGAN \cite{begusCiw}) and test how the networks classify acoustic lexical items in unobserved test data. Strong evidence in favor of lexical learning and a causal relationship between latent codes and meaningful sublexical units emerge. The architecture that combines the production and perception principles is thus able to learn to decode unique information from raw acoustic data without accessing real training data directly. We propose a technique to explore lexical (holistic) and sublexical (featural) learned representations in the classifier network. The results bear implications for unsupervised speech technology, as well as for unsupervised semantic modeling as language models increasingly bypass text and operate from raw acoustics.

\end{abstract}

\section{Introduction}

Speech technology has traditionally been divided into automated speech recognition (ASR) and speech synthesis. Hearing humans, however, perform both tasks --- speech production and speech perception --- with a high degree of mutual influence (the so-called production-perception loop; \cite{vihman15}). 

This paper proposes that the two principles should be modeled simultaneously and argues that a GAN-based model called ciwGAN/fiwGAN \cite{begusCiw} learns linguistically meaningful representations for both production and perception. In fact, lexical learning in the architecture emerges precisely from the requirement that the network for production and the network for perception interact and generate  data that is mutually informative. We show that with only the requirement to produce informative data, the models not only produce desired outputs (as argued in \cite{begusCiw}), but also learn to classify lexical items in a fully unsupervised way from raw unlabeled speech.

\subsection{Prior work}

Most of the existing models of lexical learning focus primarily on either ASR/speech-to-text (perception) or text-to-speech/speech synthesis (production; see \cite{wali22} for an overview). Variational Autoencoders (VAEs) involve both an encoder and decoder, which allows unsupervised acoustic word embedding as well as generation of speech, but these proposals only use VAEs for either unsupervised ASR \cite{chung16,chorowski19,baevski20,niekerk20} or for speech synthesis/transformation (e.g.~\cite{hsu17}). Earlier neural models replicate brain mechanisms behind perception and production \cite{guenther11}, but they do not focus on lexical learning or classification and do not include recent progress in performance of deep learning architectures.  GAN-based synthesizers are mostly supervised  and get text or acoustic features in their input \cite{kumar19,kong20,binkowski20,cong21}. \cite{donahue19} propose a WaveGAN architecture, which can generate any audio in an unsupervised manner, but does not involve a lexical classifier --- only the Generator and the Discriminator, which means the model only captures synthesis and not classification (the same is true for Parallel WaveGAN; \cite{yamamoto20}). \cite{begusCiw} proposes the first textless fully unsupervised GAN-based model for lexical representation learning, but evaluates only the synthesis (production) aspect of their model by only evaluating outputs of the Generator network. 

\subsection{New challenges}

Here, we model lexical learning with a classifier network (the Q-network) that mimics perception and lexical learning and is, crucially, trained from another network's production data (the Generator network). Using this architecture, we can \textit{both} generate new words in a controlled causal manner by manipulating the Generator's latent space \textit{as well as} classify novel words from unobserved test data with a classifier that never directly accesses the training data.

This paper also introduces some crucial new challenges to the unsupervised acoustic word embedding and word recognition paradigm \cite{dunbar17}. First, the architecture enables extremely reduced vector representations of lexical items. In fiwGAN, the network needs to represent $2^n$ classes with only $n$ variables. To our knowledge, no other proposal features such dense representation of acoustic lexical items. Second, the models introduce a challenge to learn meaningful representations of words without ever directly accessing training data. The lexical classifier network is twice removed from training data. The Q-network learns to classify words only from the Generator's outputs and never accesses training data directly. But the  Generator never accesses the training data directly either --- it learns to produce words only by maximizing the Discriminator's error rate. 

Why are these challenges important? First, representation learning with highly reduced vectors is more interpretable and allows us to analyze the causal effect between individual latent variables and linguistically meaningful units in the output of the synthesis/production part of the model (Section \ref{flip}). We can also examine the causal effect between linguistically meaningful units in the classifier's input and the classifier's output in the perception/recognition part of the model (Section \ref{flr}). 

Reduced vectors also enable analysis of the interaction between individual latent variables. For example, each element (bit) in a binary code (e.g., [1, 0], [0, 1], [1, 1]) can be analyzed as a feature $\phi_n$ (e.g.~[$\phi_1$, $\phi_2$]). Such encoding allows both holistic representation learning and featural representation learning. We can test whether each unique code corresponds to unique lexical semantics and how individual features in binary codes ([$\phi_1$, $\phi_2$]) interact/represent sublexical information (e.g.~the presence of a phoneme; Section \ref{sublexical}).

Second, humans acquire speech production and perception with a high degree of mutual influence \cite{vihman15}. 
Modeling production (synthesis) and perception (recognition) simultaneously will help us build more dynamic and adaptive systems of human speech communication that are closer to reality than current models which treat the two components separately.

Third, the paper tests learning of linguistically meaningful representations in one of the most  challenging training settings. Results from such experiments test the limits of deep learning architectures for speech processing. 

Fourth, unsupervised ASR  \cite{baevski21} and ``textless NLP'' \cite{lakhotia21} have the potential to enable speech technology in a number of languages that feature  rich phonological systems. Most deep generative models for unsupervised learning focus exclusively on either lexical (see above) or phonetic learning \cite{eloff19,shain19} and do not model phonological learning. Exploring how the two levels interact will be increasingly important as speech technology becomes available in languages other than English.

Finally, speech technology is shifting towards unsupervised learning \cite{baevski21}. Our understanding of how biases in data are encoded in unsupervised models is even more poorly understood than in supervised models. The paper proposes a way to test how linguistically meaningful units self-emerge in fully unsupervised models for word learning. 
Speech carries a lot of potentially harmful social information \cite{holliday21}; a better understanding of how linguistically meaningful units self-emerge and get encoded and how they interact with other features in the data is the first step towards mitigating the risks of unsupervised deep generative ASR models.

\begin{figure}\centering
\includegraphics[width=.5\textwidth]{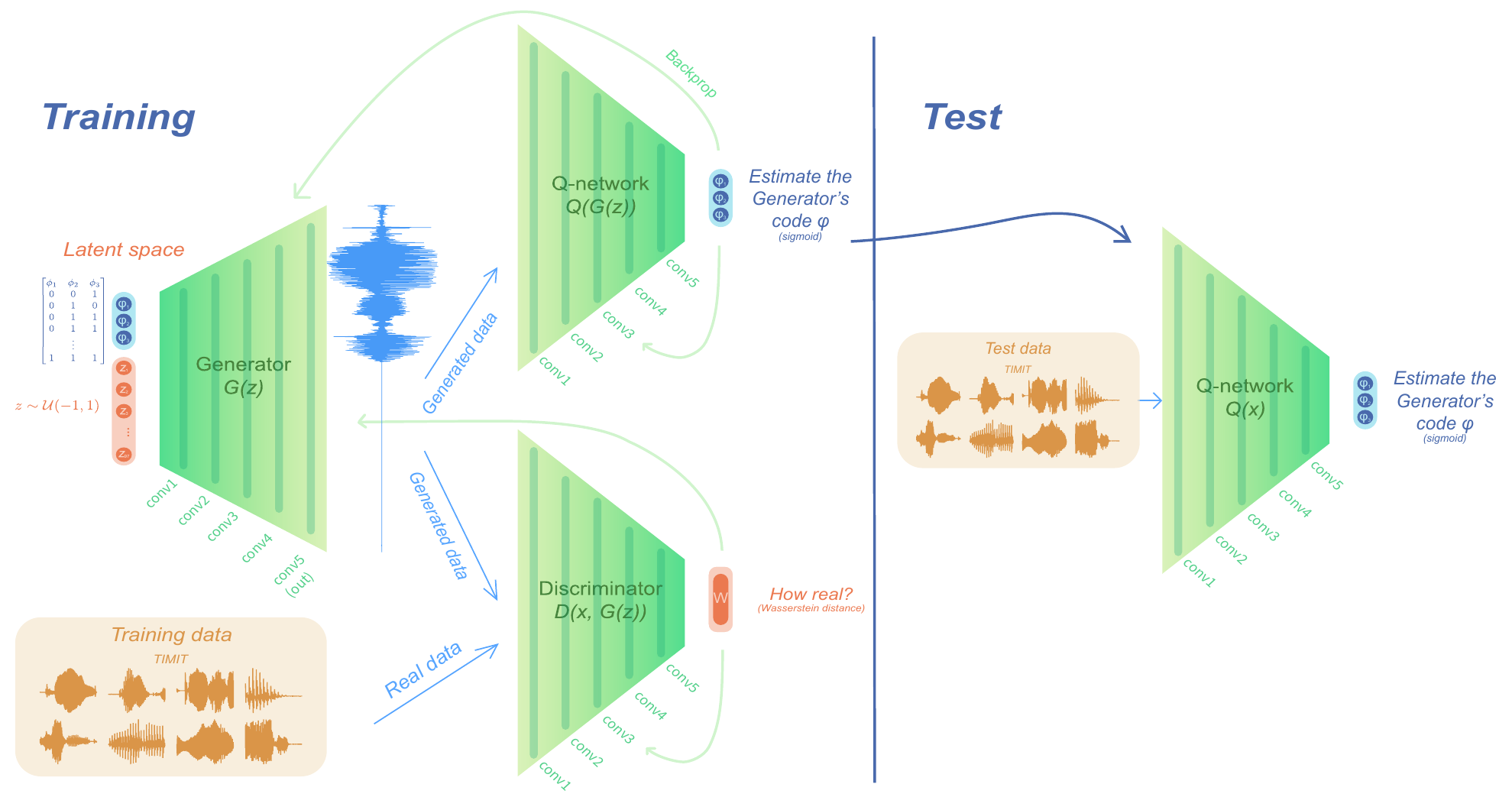}
\caption{The fiwGAN network \cite{begusCiw} in training and test tasks.}
\label{architecture}
\end{figure}

\section{Models}
We use Categorical InfoWaveGAN (ciwGAN) and Featural InfoWaveGAN (fiwGAN) architectures (\cite{begusCiw}; based on WaveGAN in \cite{donahue19} and InfoGAN in \cite{chen16}). In short, the ciwGAN/fiwGAN models each contain three networks: a Generator $G$ that upsamples from random noise $z$ and a latent code $c$ to audio data using 1D transpose convolutions, a Discriminator $D$ that facilitates the minimization of the Wasserstein distance between the distribution of the generated outputs $G(z, c)$ and real outputs $x$ using traditional 1D convolutions, and a Q-network $Q$ mirroring the Discriminator architecture that aims to recover $c$ given generated output $G(z, c)$. As in the traditional GAN framework, the Generator and Discriminator operate on the same loss in a zero-sum game, forcing the Generator to create outputs similar to the training data. However, the Generator (along with the Q-network) is additionally trained to minimize the negative log-likelihood of the Q-network, forcing the Generator to maximize the mutual information between the latent code $c$ and generated output $G(z, c)$ and the Q-network to recover the relationship between $c$ and $G(z, c)$. CiwGAN models $c$ as a one-hot vector of several classes, while fiwGAN models $c$ as a vector using a binary encoding.

Previous work  on ciwGAN and fiwGAN \cite{begusCiw} has focused on the ability of the Generator to learn meaningful representations in $c$ that encodes phonological processes and lexical learning, with no exploration of the Q-network. In this paper, we focus on the Q-network's propensity for lexical learning. Towards this end, we maintain the architecture of a separate Q-network (in contrast to the original InfoGAN proposal, where $Q$ is estimated by appending additional hidden layers after the convolutional layers of the Discriminator). This allows us to simultaneously model speech recognition using the Q-network and speech synthesis using the Generator.  

\section{Experiments}

We train three networks: one using the one-hot (ciwGAN) architecture on 8 lexical items from TIMIT, one with the binary code (fiwGAN) architecture on 8 lexical items from TIMIT. To test how the proposed architecture scales up to larger corpora, we also train a fiwGAN network on 508 lexical items from LibriSpeech \cite{librispeech}.\footnote{Checkpoints and data: \url{doi.org/10.17605/OSF.IO/NQU5W}.\label{fn}}

\subsection{Data}

The lexical items used in 8-words models are: \textit{ask}, \textit{dark}, \textit{greasy}, \textit{oily}, \textit{rag}, \textit{year}, \textit{wash}, and \textit{water}. A total of 4,052 tokens are used in training (approximately 500 per each word). The words were sliced from TIMIT and padded with silence into 1.024s  .wav files with 16kHz sampling rate which the Discriminator takes as its input.

In the LibriSpeech experiment, 508 words were chosen. We discarded the 78 most common lexical items in the LibriSpeech train-clean-360 dataset \cite{librispeech} because of their disproportionate high frequency (5,290 to 224,173 tokens per word). We then arbitrarily choose the 508 next most common words for training, resulting in a total of 757,120 tokens. The individual counts for each word in the training set ranges from 571 to 5,113 tokens.

\subsection{Perception/classification}

To test if the Q-network is successful in learning to classify lexical items without ever accessing training data, we take the trained Q-network from the architecture (in Figure \ref{architecture}) and feed it novel, unobserved data. In other words, we test if the Q-network can correctly classify novel lexical items by assigning each lexical item a unique code.

Altogether 1,067 test data in raw waveforms from unobserved TIMIT were fed to the Q-network (both in the ciwGAN and fiwGAN architectures). The raw output of this experiment are pairs of words with their TIMIT transcription and the unique code that the Q-network outputs in its final layer. We test the performance of the models using inferential statistics rather than comparison to existing models due to the lack of models with similarly challenging learning objectives.

\begin{figure}\centering
\includegraphics[width=.27\textwidth]{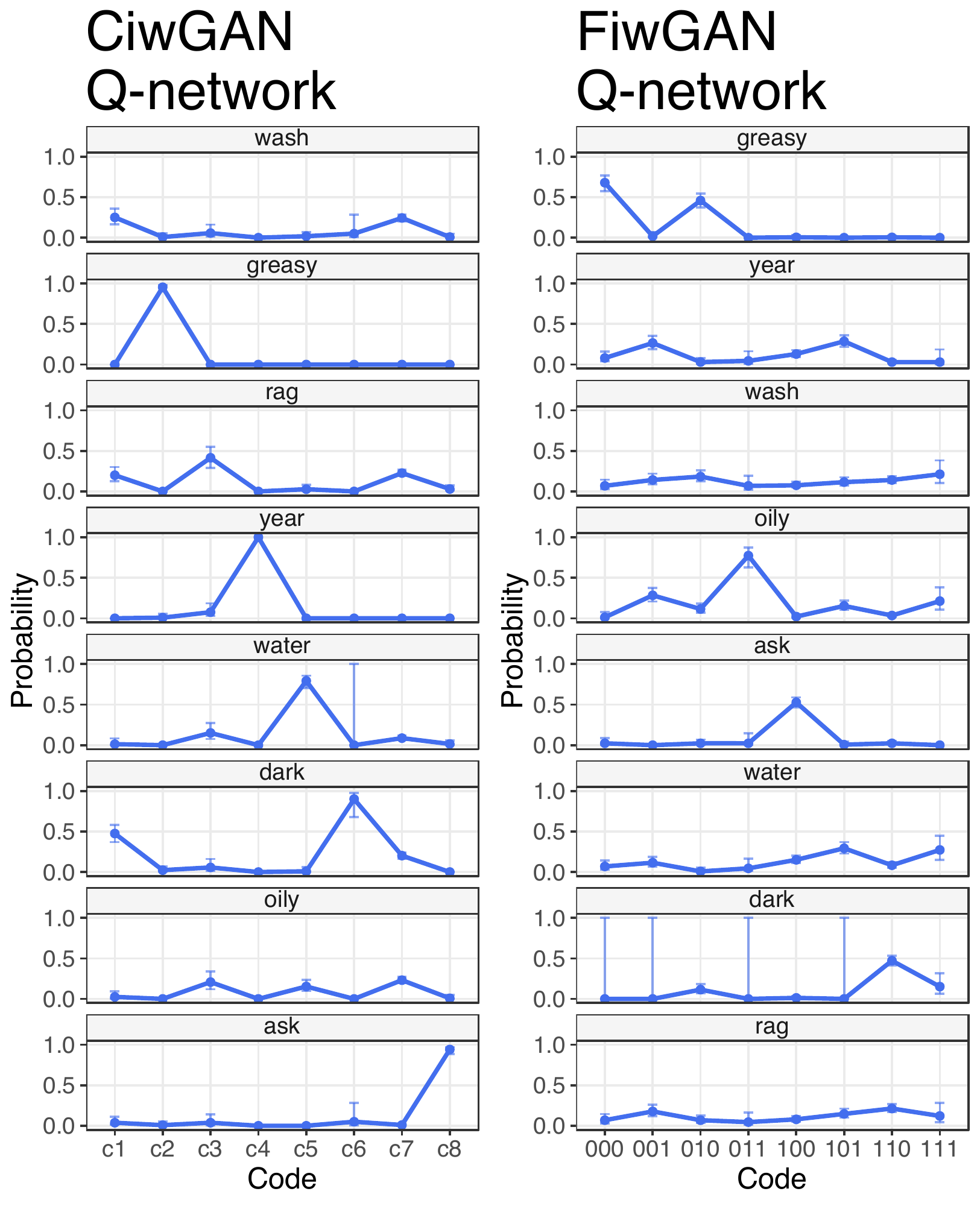}
\caption{Estimates of a multinomial regression model.}
\label{fiwganFigure}
\end{figure}

To perform hypothesis testing on whether lexical learning emerges in the Q-network, we fit the word/code pairs to a multinomial logistic regression model with the \textit{nnet} package \cite{nnet}. 
In the ciwGAN setting (one-hot encoding), AIC of a model with $c$ as a predictor is substantially lower ($2129.1, df=56$) than the empty model  ($4448.2, df=7$).  Figure \ref{fiwganFigure} gives predicted values for each code/word. The figure suggests that most words (with some exceptions especially in the fiwGAN model)  have a clear and substantial rise in estimates for a single unique code. This suggests that the Q-network learns to classify novel unobserved TIMIT words into classes that correspond to lexical items.

Lexical learning emerges in the binary encoding (fiwGAN) as well, but the code vector is even more reduced in this architecture (3 variables total), which makes error rates higher compared to the ciwGAN architecture (Figure \ref{fiwganFigure}). 

\subsection{Production/synthesis}

To test the production (synthesis) aspect of the model, we generate 100 outputs for each unique latent code $c$  both in the ciwGAN and fiwGAN setting (1,600 outputs total). According to \cite{begus19,begus2020identity,begusCiw}, setting latent codes to marginal values outside of the training range while keeping the rest of the latent space constant reveals the underlying value of each latent code, which is why we generate data with code variables set at 3 (e.g.~[0, 0, 3], [0, 3, 3], etc). One hundred outputs per each code for each model (ciwGAN and fiwGAN) were analyzed by a compensated trained phonetician who was not a co-author on this paper. The annotator annotated generated outputs as either featuring the eight lexical items, deviating from the eight items (annotated as \textit{else}), or as unintelligible outputs (also \textit{else}). For coding of annotations, see online data in fn.~\ref{fn}.

Code variables are significant predictors of generated words according to the AIC test in both models. The learned representations are very similar and mostly match across the Q-network and in the Generator. One advantage of the Generator network is that we can force categorical or near categorical outputs by manipulating latent variables to marginal values outside of training range (e.g.~in our case to 3). For example, \textit{greasy} has 100\% success rate in ciwGAN; \textit{water} 99\% in fiwGAN and 96\% in ciwGAN.

\section{Holistic and featural learning}
\label{sublexical}

Binary encoding allows simultaneous holistic encoding of lexical semantic information (unique code = lexical item) as well as featural learning, where features (bits) correspond to sublexical units such as phonemes (e.g.~[s] or [\textipa{S}]). This paper proposes a technique to explore lexical and sublexical learned representations in a classifier network.  To test whether evidence for sublexical learning emerges in the perception aspect of the proposed model, we annotate inputs to the Q-network for any sublexical property and use regression analysis with each feature (bit) as a predictor to test how individual features correspond to that property.

\subsection{TIMIT}

We focus on one of the the most phonetically salient sublexical properties in the training data: presence of a fricative [s], [\textipa{S}]. We include the word for \textit{dark} among the words containing [s] because a high proportion of \textit{dark} tokens feature [s] frication (due to \textit{dark} standing before \textit{suit} in TIMIT). The data were fit to a logistic regression linear model with presence of [s] in the input test data as the dependent variable and the three features   ($\phi_1$, $\phi_2$, $\phi_3$) as predictors. Estimates of the regression model  suggest that the network encodes a sublexical phonemic property (presence of frication noise of [s]) with $\phi_3 = 0$ ($\beta=-0.5,z= -2.9, p=0.004$ for $\phi_1$, $\beta=   0.2,z= 1.0,p= 0.3$, for $\phi_2$ and $\beta=  -2.5,z= -13.8,p<0.0001$ for $\phi_3$).

\subsection{LibriSpeech}

To test how the proposed technique of unsupervised lexical and sublexical learning extends to larger corpora, we test the Q-network trained on 508  lexical items from LibriSpeech. The model has 9 latent feature variables $\phi$ which yields $2^9=512$ classes. Altogether 10,914 test tokens (withheld from training) of the 508 unique words were fed to the Q-network in fiwGAN architecture trained for 61,707 steps. 

\subsubsection{Holistic representation learning}
\begin{figure}[t]\centering
\includegraphics[width=.3\textwidth]{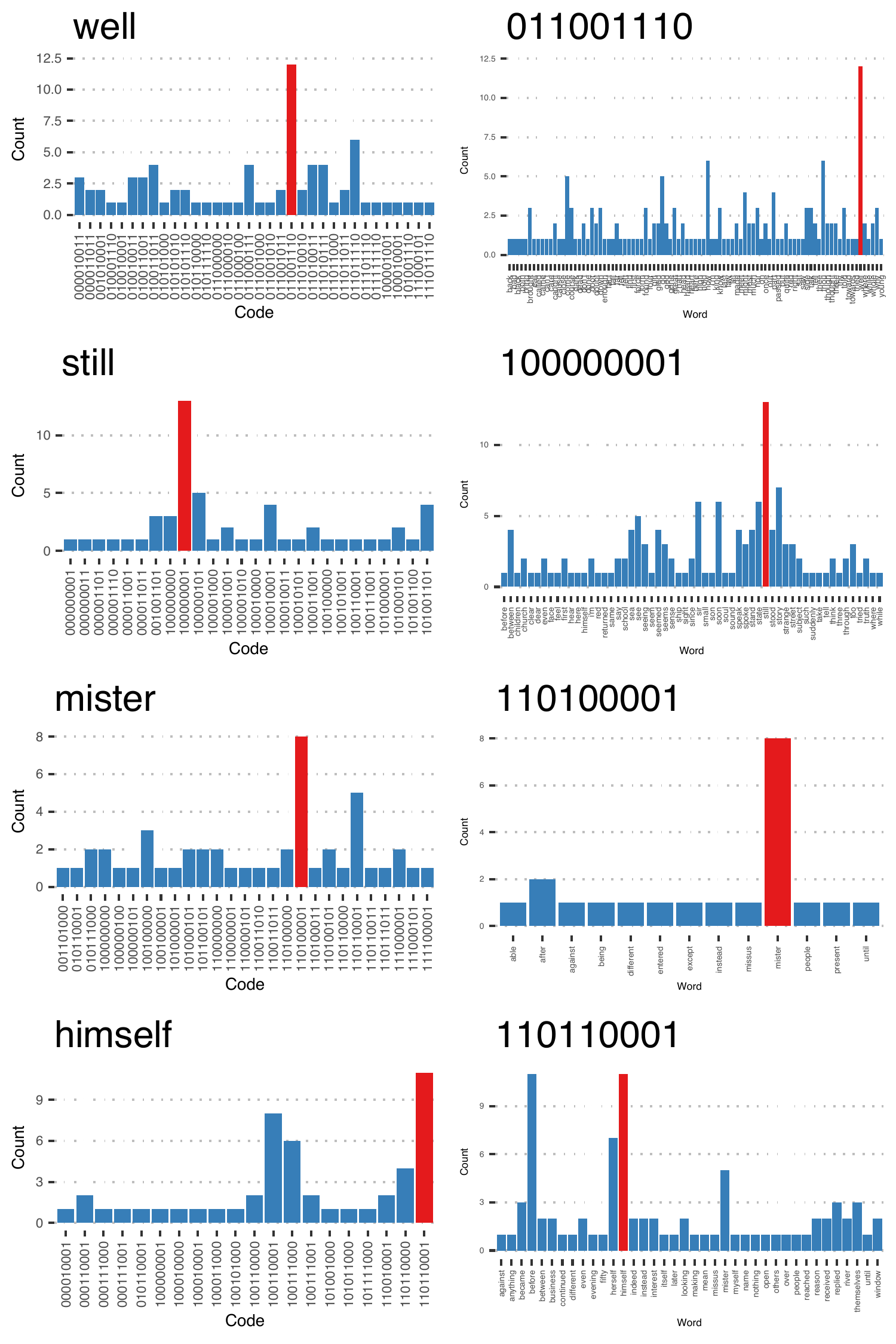}
\includegraphics[width=.3\textwidth]{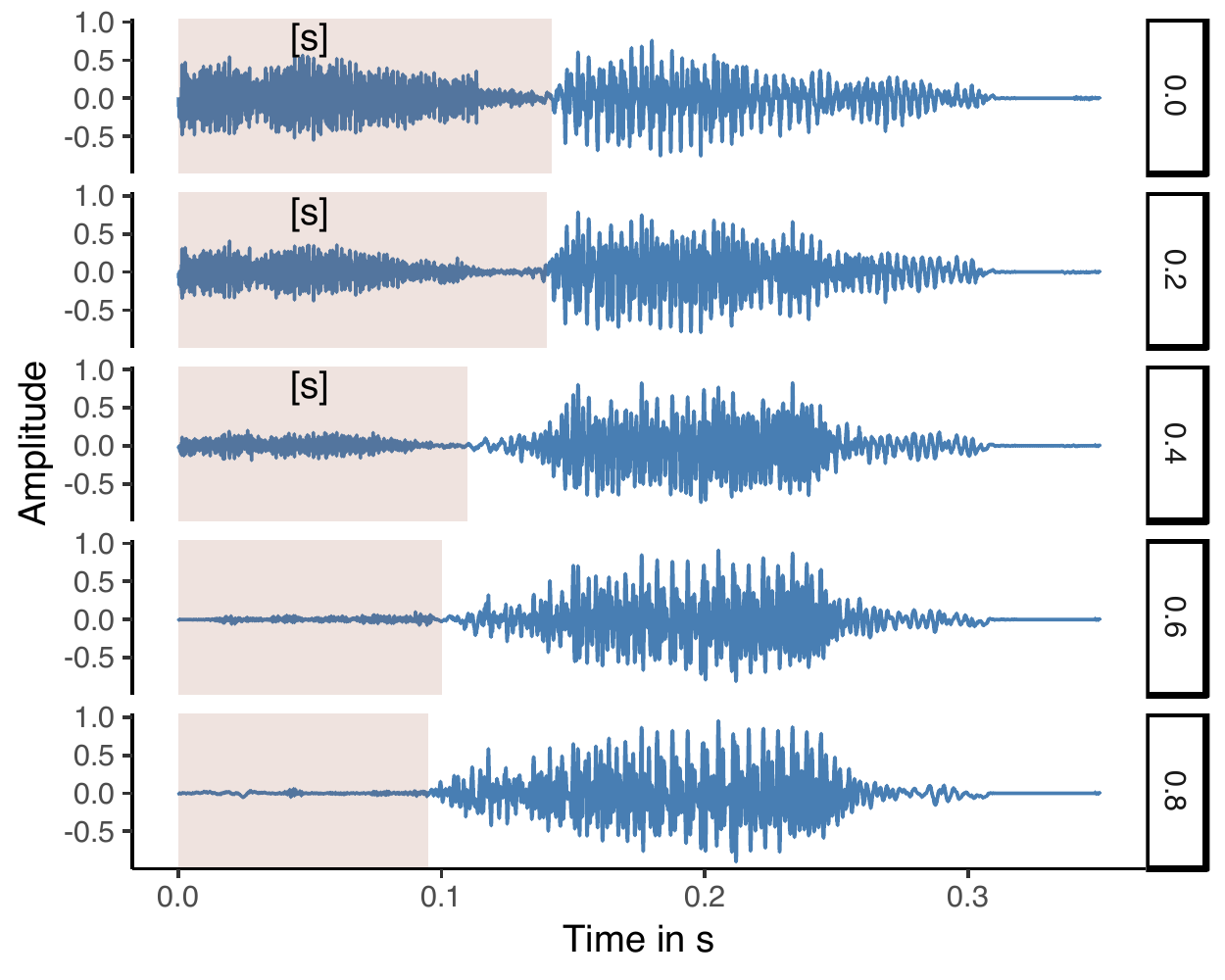}
\caption{(top left) Raw counts of code distribution per each of the four chosen tested words (from unobserved test data). The code with highest count is color-coded in red. (top right) Raw counts of all words classified with the code that has the highest count for each word from the left graph. Words that were never classified with this code are not on the graph. The word with highest count is color-coded in red. (bottom) 
Outputs of the Generator network (waveforms) when $\phi_2$, $\phi_3$, and $\phi_5$ are simultaneously interpolated from 0.0 to 0.8 while all other latent variables are held constant. }
\label{wellstill}
\end{figure}

First, raw classification of outputs suggest that holistic lexical learning in the Q-network emerges even when the training data contains a substantially larger set (508 items and a total of 757,120 tokens) and a more diverse corpus. The training data here too is twice removed from the Q-network and the test data was never part of the training. Figure \ref{wellstill} illustrates four chosen words and the codes with which they are represented. Each word features a peak in one unique code. To verify that this particular code indeed represents that particular word, we also analyze which other words are classified with the most frequent code for each of the four chosen word. There too, each code represents one word more strongly. 

To test how common such well-learned representations are, we randomly selected 20 out of the 508 words from LibriSpeech, which includes words that occur extremely infrequently (e.g.~$\mathrm{N}=7$) in both the train and test sets. Of the 20 randomly selected words, 4 (20\%) have representations where the code most frequently assigned to a word (peak in Figure \ref{wellstill} top left) also has the highest count (peak in Figure \ref{wellstill} top right) of that same word when compared to all words labeled with that code (e.g.~011001110 is most common code given to \textit{well}, and \textit{well} is the most common word that is labelled as 011001110; Figure \ref{wellstill}).
 In 5 further cases (25\%), two or more peaks have the same, but not higher counts than the word/code peak pair (for a total of 45\% of successful outcomes if both groups are counted as successful). In the remaining 55\% (11 items), the peaks do not match across the word/code pairs. We counted one case with all counts equal across the word/code pair as unsuccessful.

These counts are fully deterministic and therefore conservative. The distribution of code variables per each word are, however, not independent. For example, the second most common code for \textit{mister} in Figure \ref{wellstill} differs from the most common one in only one feature (bit). Violation in a single feature value is equally treated as violation in multiple feature values in our counts. 
Likewise, there is substantial amount of phonetic similarity in words classified by a single code. For example, the word most commonly classified with [100000001] is indeed \textit{still}, but other frequent words for this classification code are \textit{state}, \textit{stand}, \textit{stood}, \textit{story}, etc. (Figure \ref{wellstill}).

\subsubsection{Featural representation learning}
\label{flr}

These similarities suggest that the network encodes sublexical properties using individual features in the binary code. To quantitatively test this hypothesis, we test how the network encodes presence of word-initial [\#s]. Frication noise of [s] is a phonetically salient property and restricting it to word-initial position allows us to test featural and positional (temporal) encoding.

Librispeech word/Q-network code pairs are annotated for presence of word-initial [s] (dependent variable) and fit to a logistic regression linear model with the nine feature variables $\phi_{1-9}$ (bits) as independent predictors. 
Three features ($\phi_2$, $\phi_3$, and $\phi_5$) correspond to presence of initial [\#s] substantially more strongly than other features. It is reasonable to assume that the network encodes this sublexical contrast with the value of the three features ($\phi_2$, $\phi_3$, $\phi_5$) at 0. It would be efficient if the network encodes word-initial [\#s] with 3 features, because there are approximately 54 s-initial words. The 6 feature codes remaining besides $\phi_2$, $\phi_3$, $\phi_5$  allows for $2^6=64$ classes. 

To verify this hypothesis, the presence of [\#s] in input words  (dependent variable) is fit to a logistic regression model with only one predictor: the value of the three features $\phi_2$, $\phi_3$, $\phi_5$ with two levels: 0 and 1. Only 5.0\% [4.6\%, 5.5\%] of words classified with $\phi_2$, $\phi_3$, and $\phi_5$ = 1  contain word-initial [\#s], while 47.9\% [44.3\%, 51.5\%] of words classified as $\phi_2$, $\phi_3$, and $\phi_5$ = 0 contain word-initial [\#s].

\subsection{Featural learning in production}
\label{flip}

The fiwGAN architecture allows us to test both holistic and featural learning in both production and perception. Value 0 for $\phi_2$, $\phi_3$, $\phi_5$ has been associated with word-initial [\#s] in the Q-network (perception). To test whether the Generator matches the Q-network in this sublexical encoding, we generate sets of outputs in which all other $\phi$ variables (except $\phi_2$, $\phi_3$, and $\phi_5$) and all $z$-variables are held constant, but the $\phi_2$, $\phi_3$, and $\phi_5$ variables are interpolated from 0 to 3 in intervals of 0.2. We analyze 20 such outputs (where the other $\phi$ variables and $z$-variables are sampled randomly for each of the 20 sets).

In 11 out of the 20 generated sets (or 55\%), word-initial [\#s] appears in the output for code $\phi_2$, $\phi_3$, $\phi_5$ = 0 and then disappears from the output as the value is interpolated (annotated by the authors because presence of [s] is a highly salient feature).  Additionally, in the majority of these cases (approximately 8), the change from [\#s] to some other word-initial consonant is the only major change that happens as the output transitions from [s] to no [s] with interpolation. In other words, as we interpolate values of the three features representing [\#s], we observe a causal effect in the generated outputs as [\#s] gradually changes into a different consonant with other major acoustic properties remaining the same in the majority of cases. Figure \ref{wellstill} illustrates this causal effect: the amplitude of the frication noise of [\#s] gradually attenuates with interpolation, while other acoustic properties remain largely unchanged. The sublexical encoding of word-initial [\#s] is thus causally represented with the same code both in the Generator network and in the Q-network. 
\section{Conclusion}

This paper demonstrates that a deep neural architecture that simultaneously models the production/synthesis and perception/classification learns linguistically meaningful units --- lexical items and sublexical properties --- from raw acoustic data in a fully unsupervised manner. 
 We also argue that we can simultaneously model holistic lexical representation learning (in the form of unique binary codes) and sublexical (phonetic and phonological) representations in the form of individual feature codes (bits) in the fiwGAN architecture.

\bibliographystyle{IEEEtran}

\bibliography{begusGANbib.bib}
\end{document}